\documentclass{article}

\PassOptionsToPackage{numbers,square}{natbib}
\usepackage[preprint]{neurips_2026}
\usepackage{graphicx}
\usepackage{multirow}
\usepackage{graphicx}
\usepackage{makecell}
\usepackage{subcaption}
\usepackage{tabularx}
\usepackage{array}
\usepackage{wrapfig}
\usepackage{algorithm}
\usepackage{algpseudocode}
\usepackage{picinpar}
\usepackage{enumitem}
\usepackage{cutwin}
\usepackage[table]{xcolor}
\usepackage{booktabs}
\usepackage{float}
\usepackage[utf8]{inputenc} 
\usepackage[T1]{fontenc}    
\usepackage{authblk}
\usepackage{hyperref}       
\usepackage{url}            
\usepackage{booktabs}       
\usepackage{amsfonts}       
\usepackage{nicefrac}       
\usepackage{amsmath}
\usepackage{microtype}      
\usepackage{xcolor}         
\usepackage{amsmath}
\usepackage{booktabs}
\usepackage{algorithm}
\usepackage{algpseudocode}
\usepackage{authblk}

\title{Towards a Dynamic and Fixed-budget Memory Bank for Efficient Streaming Video Understanding}

\author{
\textbf{Baiyang Song$^{1}$ \quad Yuli Lin$^{1}$ \quad Qiong Wu$^{1}$ \quad Tao Chen$^{1}$ \quad Jun Peng$^{1}$\quad Xiao Chen$^{1}$\quad Yiyi~Zhou$^{1\dagger}$ \quad Rongrong Ji$^{1}$}\\
\textsuperscript{1}Key Laboratory of Multimedia Trusted Perception and Efficient Computing,\\
Ministry of Education of China, Xiamen University, 361005, P.R. China\\
\texttt{\{songbaiyang, linyuli330, qiong, taochen\}@stu.xmu.edu.cn \quad pengjun@outlook.com \quad 2912@mju.edu.cn \quad \{zhouyiyi, rrji\}@xmu.edu.cn \quad}
}

\begin{document}

\maketitle
\renewcommand{\thefootnote}{}
\footnotetext{$^{\dagger}$ Corresponding author.}

\begin{abstract}
\vspace{-0.1cm}
Currently, streaming video understanding is still a daunting task for existing \emph{multimodal large language models} (MLLMs). 
Its difficulties not only lie in handling the ever-increasing video frames, but also in the unpredictability of future video content and input instructions.
In this paper, we study this task from the perspective of constructing a dynamic but fixed-budget memory bank, and propose a novel and training-free approach termed \emph{\textbf{CausalMem}}. 
CausalMem is dedicated to constructing a dynamic visual memory update mechanism, thereby maximizing the amount of information in streaming video within a limited memory space, much like the human brain. In practice,
CausalMem estimates the redundancy of visual tokens and updates the memory bank via an online semantic basis, which models the principal semantics of the observed video stream. 
To validate CausalMem, we apply it to two representative MLLMs, namely LLaVA-OneVision and Qwen2.5-VL respectively, and conduct
extensive experiments on both streaming and offline video 
understanding benchmarks. 
The experimental
results not only show the great advantages than existing methods under both streaming and offline settings, \emph{e.g.}, $+3.2\%$ and $+3.0\%$ average accuracy gains respectively, but also witness the superior semantic preservation for streaming videos, \emph{e.g.}, using 12$k$ token budgets to memorize hour-long streaming videos, which achieves more than \textbf{20$\times$} visual token compression ratio and only occupies about \textbf{82 MB} storage. 
\textbf{Our code} is given in \href{https://github.com/hktk07/CausalMem}{CausalMem}.

\end{abstract}

\vspace{-0.4cm}
\section{Introduction}
\vspace{-0.1cm}

After achieving great successes in a variety of image-language tasks \cite{antol2015vqa,chen2015microsoft,singh2019towards,mishra2019ocr}, recent \emph{multimodal large language models} (MLLMs) \cite{tong2025flashsloth,liu2023visual,li2023blip} also yield notable progress in the video domain \cite{zhang2024llava,li2024llava,bai2025qwen25vltechnicalreport}. 
To overcome the substantial computation caused by the excessive video tokens, numerous works \cite{ye2025fit,ju2026forestprune,shao2025holitom,fan2026flashvid,jiang2025kind} have been recently devoted to the research of efficient video understanding, employing techniques such as 
\emph{token pruning}~\cite{yang2025visionzip,ye2025fit,ju2026forestprune} and \emph{KV Cache compression}~\cite{yang2025streammem,di2025streaming,chen2026flexmem}.
In terms of video content compression, current research primarily focuses on offline video processing, which typically requires globally semantical compression strategies based on the entire video content.
Moreover, for streaming video understanding, the fixed-budget compression still remains unexplored.

To this end, efficient streaming video understanding is still an intractable problem for existing MLLMs \cite{li2024llava,zhang2024llava,bai2025qwen25vltechnicalreport}, of which difficulties mainly lie in two aspects. 
Firstly, MLLMs need to process the video of potentially unbounded length in the streaming scenario. 
For instance, LLaVA-OneVision \cite{li2024llava} needs to cope with more than 700$k$ visual tokens for a two-hour video (1 FPS). 
Moreover, even the advanced video compression methods \cite{chen2025towards,ju2026forestprune} will encounter a rapid accumulation of visual tokens over time, placing a heavy burden on both computation and storage. 
Secondly, when compressing streaming videos, it is almost impossible to predict future video content and potential user questions.
This case renders many existing compression methods unapplicable, such as those based on global redundancy \cite{ju2026forestprune,shao2025holitom} or text-guided estimation \cite{zhang2026one,li2025less}.
Therefore, we raise a question that

\emph{
"Is it possible to achieve the dynamic and budgeted visual memory bank for efficient streaming video understanding, much like the human brain?" }


\begin{figure*}[t]
    \centering
    \includegraphics[width=0.95\columnwidth]{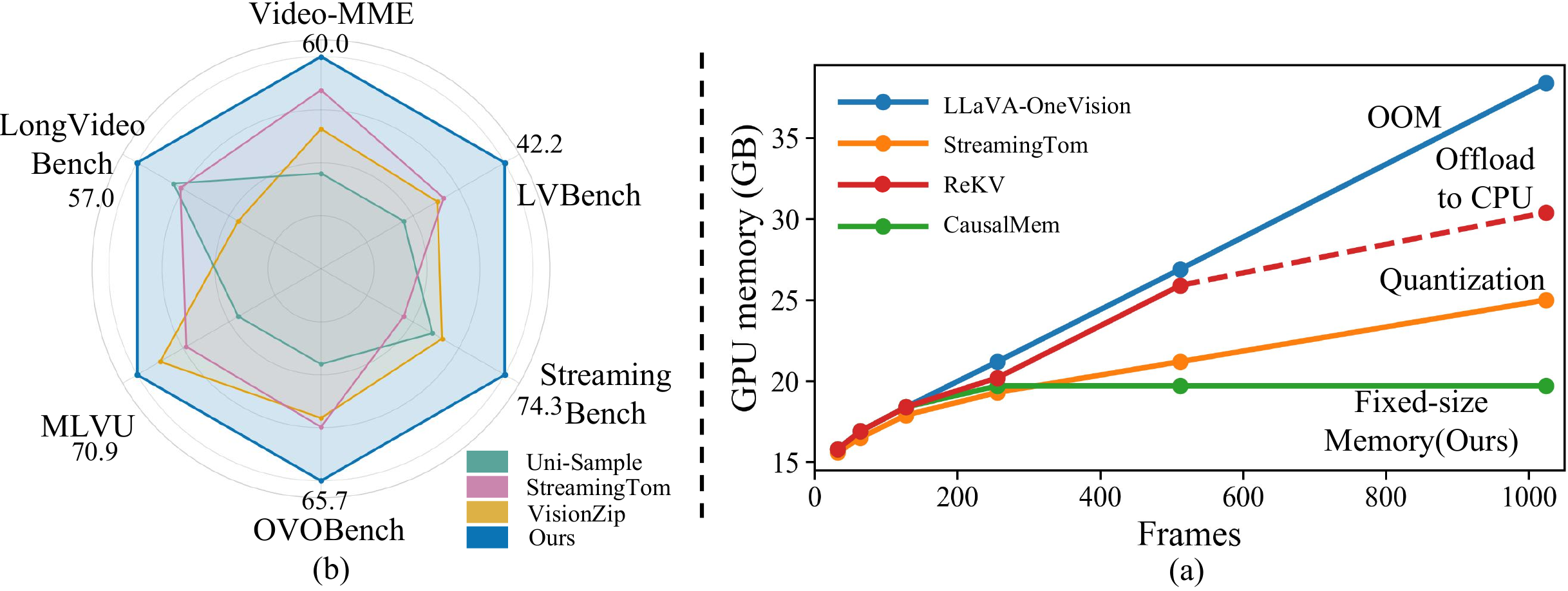}
    \vspace{-0.15cm}
    \caption{(a) CausalMem consistently outperforms existing visual compression methods across streaming and offline video understanding benchmarks. The base MLLM used is LLaVA-OneVision. (b) Compared with existing streaming video methods, CausalMem maintains a fixed-size memory bank, enabling streaming video understanding with bounded computation and storage overhead.}
    \label{fig:overall}
    \vspace{-0.5cm}
\end{figure*}


In this paper, we study this problem from the perspective of constructing a dynamic and fixed-budget visual memory bank, and propose a novel and training-free method for existing MLLMs, termed \textbf{CausalMem}. 
To relax the burdens of both computation and storage, CausalMem will maintain a fixed-size memory bank for streaming video tokens, while using a memory update mechanism to meet the properties of streaming video understanding. 
In practice,  CausalMem draws on the principle of \emph{Information Maximization} \cite{deerwester1990indexing} and builds a compact but evolving \emph{semantic basis} for the memory bank, thereby well estimating the visual tokens for memory update. 
In this way, CausalMem can not only well reduce the visual redundancy but also maximize the information gain of the  memory bank.

To validate CausalMem, we apply it to two  MLLMs, namely LLaVA-OneVision~\cite{li2024llava} and Qwen2.5-VL~\cite{bai2025qwen25vltechnicalreport}, and conduct
extensive experiments on both streaming and offline video understanding
benchmarks. The experimental results show that CausalMem not only achieves
obvious performance gains than the compared streaming or offline methods, \emph{e.g.}, $+3.2\%$ and $+3.0\%$ average performance gains
on streaming and offline benchmarks respectively,
but also preserve the principal semantics of streaming videos by only using 12$k$ token budgets, \emph{e.g.}, achieving more than $20 \times$ 
visual token compression ratio for an hour-long video
while consuming the storage of only 82MB.

Overall, our contributions are twofold:
\begin{itemize}[leftmargin=*, itemsep=0pt, topsep=2pt]
    \item We study the efficient streaming video understanding from the perspective of a dynamic yet fixed budget memory bank under strict causality, and propose a novel and training-free approach for existing MLLMs, termed CausalMem. 
    \item The proposed CausalMem obtains competitive or even better performance than existing efficient video compression methods on both streaming and offline benchmarks, and also exhibits its superiority in streaming video semantic compression with fixed token budgets.
\end{itemize}


\vspace{-0.3cm}
\section{Related Work}
\vspace{-0.2cm}
\textbf{Offline video understanding.} 
Recent Multimodal Large Language Models~\cite{liu2023visual,bai2025qwen25vltechnicalreport,team2024gemini} have shown strong capabilities in offline video understanding. 
However, as video length increases, the number of visual tokens increases linearly, leading to prohibitive computational cost and memory overhead.
Existing methods address this issue from two main perspectives. 
One is adopting Retrieval-Augmented Generation (RAG), in which a subset of task-relevant frames is selected based on vision–text similarity, thereby reducing the number of frames processed by the MLLM. 
While effective, they rely on the availability of user queries in advance, limiting their applicability to online settings.
The other one focuses on visual token compression. 
They reduce the token sequence length via token pruning or merging~\cite{ju2026forestprune, shao2025holitom, bolya2022token, shang2025llava}, either based on global video content or query-aware relevance. 
Despite their effectiveness, they require whole video access and are thus unsuitable for streaming scenarios.

\textbf{Streaming Video Understanding.} 
To address the above limitations, recent works focus on streaming video understanding, where visual tokens must be processed incrementally based on observed frames.
Most existing methods aim to control the growth of visual tokens through memory or compression mechanisms.
A representative direction is KV cache compression, where methods such as StreamMem~\cite{yang2025streammem}, ReKV~\cite{di2025streaming} and LiveVLM~\cite{ning2025livevlm} compress the key–value cache generated by visual tokens to prevent unbounded memory growth.
Another line of work introduces structured memory mechanisms. 
For example, FluxMem~\cite{xie2026fluxmem} organizes visual tokens into hierarchical memory with a First-In-First-Out (FIFO) strategy to manage short-, medium-, and long-term information.
In addition, StreamingTOM~\cite{chen2025streamingtom} combines token compression before the LLM with KV cache quantization to further reduce memory overhead.
Beyond compression, some methods also incorporate query-aware retrieval. 
For instance, LiveVLM~\cite{ning2025livevlm} and StreamingTOM~\cite{chen2025streamingtom} retrieve the most relevant visual tokens conditioned on the user query.
Despite these advances, existing methods primarily rely on local redundancy within the observed visual tokens or require query guidance, which limits their effectiveness in fully capturing the global semantic structure of the video stream.

\section{Method}
\subsection{Overview}

Given a video \(V\) and a question \(Q\), an MLLM aims to generate the answer by maximizing the conditional generation objective:
\begin{equation}
    p(Y \mid \mathbf{V}, Q) = \prod_{i=1}^{L} p(y_i \mid G, \mathbf{V}, Q, Y_{<i}),
\end{equation}
where $G$ denotes the generator of MLLM, $p(\cdot)$ denotes the probabilities of the predicted word,
$Y=\{y_1,...,y_L\}$ is the answer sequence and $L$ is its 
length. $\mathbf{V}$ is 
a subsequence of visual tokens encoded from the video $V$ and $Y_{<i}$ denotes the answer subsequence before the
i-$th$ step.

Due to the limited context length and memory budget of MLLMs, some methods \cite{ju2026forestprune,shao2025holitom}
 apply token compression to 
reduce the length of $\mathbf{V}$, \emph{e.g.}, by pruning or
merging the redundant tokens\cite{ju2026forestprune,shao2025holitom,fan2026flashvid}. Meanwhile, some endeavors \cite{chen2026flexmem,shu2025video} compress the KV caches of MLLMs generated by visual tokens.
However, these compression methods usually need to 
operate on the complete video or ignore the upper limit of both computation and storage, which are inferior in streaming scenarios.

To this end, we aim to maintain a visual memory bank $\mathbf{M} \in \mathbb{R}^{b\times d}$ with a fixed budget (length) $b$, $d$ is the visual token feature dimension, and generate the answer based on this memory bank:
\begin{equation}
    p(Y \mid \mathbf{V}, Q) = \prod_{i=1}^{L} p(y_i \mid G,\mathbf{M}, Q, Y_{<i})
\end{equation}
The memory bank keeps a compact set of visual tokens while being dynamically updated as 
each new frame \(I\) arrives. Therefore, the problem we study becomes how to 
maintain $\mathbf{M}$ online so that it preserves the most informative 
tokens of the observed video stream under strict causality.

To this end, we first construct and update an \emph{online semantic basis} $\mathbf{B} \in \mathbb{R}^{q\times d}$
, where $q$ is the maximum number of semantic basis vectors in $\mathbf{B}$ and $q \ll t \times N$, where \(t\) and \(N\) denote the number of streaming frames and visual tokens per frame, respectively.  

In practice, $\mathbf{B}$ is used to dynamically model the principal semantics of the observed video stream,
based on which we can estimate the redundancy of streaming visual tokens and update the memory bank.
Then we compress the visual tokens into $\mathbf{M}$ according to their redundancy once $\mathbf{M}$
achieves the upper limit. Note that, the updating process of 
$\mathbf{B}$ and $\mathbf{M}$ are both causal, \emph{i.e.} 
unpredictable to future frames and the question.
Therefore, $M$ can keep a fixed budget under strict causality.

\begin{figure*}[t]
    \centering
    \includegraphics[width=\columnwidth]{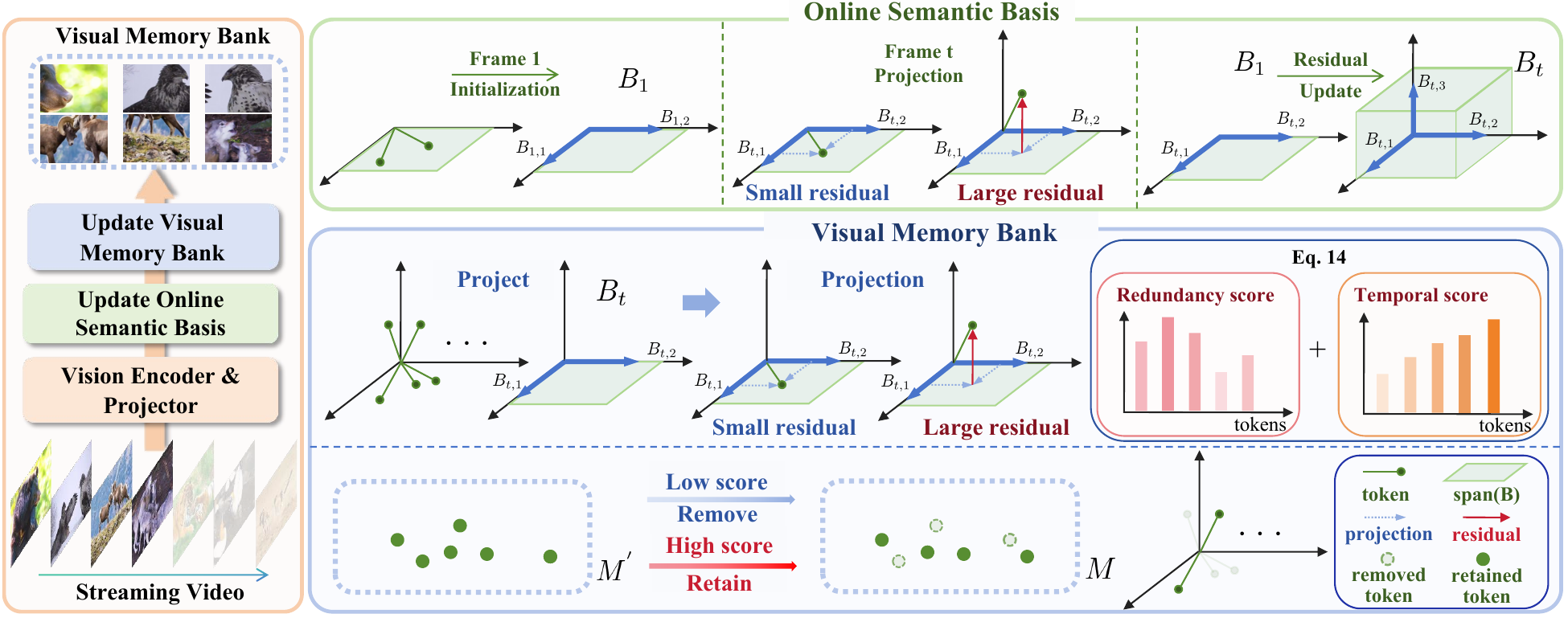}
    \vspace{-0.2cm}
    \caption{Overview of CausalMem. Given a streaming video, CausalMem encodes incoming frames and maintains a fixed-size visual memory bank under strict causality. An online semantic basis is dynamically updated to capture the principal semantics of the observed stream, and token residuals with respect to this basis are used to estimate redundancy. The memory bank is then updated by retaining tokens with low redundancy and temporal recency, thereby maximizing its information. }
    \label{fig:overall}
    \vspace{-0.5cm}
\end{figure*}




\subsection{Online Semantic Basis}

To model the principle semantics of the video stream at the $t$-th time step, we maintain an \emph{Online Semantic Basis} $\mathbf{B}_t \in \mathbb{R}^{q_t\times d}$, where $q_t \leq q$ and $q$ is the max number of basis vectors.
Let the visual tokens of the $t$-th frame be denoted as $\mathbf{X}_t \in \mathbb{R}^{N\times d}$.
At the 1$st$ step, we initialize the semantic basis from $\mathbf{X}_1$ via compact \emph{Singular Value Decomposition (SVD)} \cite{golub2013matrix}:
\begin{equation}
  \mathbf{X}_1 \approx \mathbf{U}_1 \mathbf{\Sigma}_1 \mathbf{W}_1^\top.
\end{equation}
Here, $\mathbf{U}_1 \in \mathbb{R}^{N\times q_1}$ denotes the left singular vectors, capturing the token wise variation. $\mathbf{\Sigma}_1 \in \mathbb{R}^{q_1 \times q_1}$ is the diagonal matrix of singular values quantifying the importance of each principal component, and $\mathbf{W}_1^\top \in \mathbb{R}^{q_1 \times d}$ denotes right singular vectors, whose columns represent the dominant semantic vectors in the feature space. 
We retain the top $q_1 \in (0,q]$ singular vectors and initialize the basis as
\begin{equation}
\mathbf{B_1} = \mathbf{W}_1 = [w_1, w_2, ... , w_{q_1}]\in\mathbb{R}^{q_1\times d},
\end{equation}
where $w_i \in \mathbb{R}^d$ denotes the $i$-th right singular vector.
Meanwhile, we introduce a \emph{basis activity score} $\mathbf{s}_t \in \mathbb{R}^{q_t}$ to track the usage frequency of each basis vector, which is initialized by 1 and used to guide the update of the semantic basis $\mathbf{R}$.

For the evolving video stream, the basis matrix is updated at each time step.
Given a new frame, we project its visual tokens $\mathbf{X}_t$ onto the current basis $\mathbf{B}_{t-1}$ and reconstruct them by
\begin{equation}
   \mathbf{\hat{X}}_t = \mathbf{X}_t \mathbf{B}_{t-1}^\top \mathbf{B}_{t-1},
   \label{eq:projection}
\end{equation}
where $\mathbf{\hat{X}}_t \in \mathbb{R}^{N \times d}$ represents the reconstructed token features from their projection onto the subspace spanned of the semantic basis $\mathbf{B}_{t-1}$. 
Then, the residual matrix $\mathbf{R}_t = \mathbf{X}_t - \mathbf{\hat{X}}_t$ captures the information that cannot be represented by $\mathbf{B}_{t-1}$.
In this case, we can measure the redundancy of each token $\mathbf{X}_{t,i}$ via its residual norm, defined by
\begin{equation}
  e_{t,i} = \|\mathbf{R}_{t,i}\|_2,
  \label{eq:residual}
\end{equation}
where $\|\cdot\|_2$ denotes the $L_2$ norm.
A larger $e_{t,i}$ indicates that the tokens contain more novel and informative semantics, also indicating less redundancy.

To update the basis $\mathbf{B}_{t-1}$, we select the top-$k$ least redundant tokens to form candidate tokens $\mathbf{X}_t^{\mathrm{cand}} \in \mathbb{R}^{k\times d}$, whose residual $\mathbf{R}_t^{\mathrm{cand}} \in \mathbb{R}^{k\times d}$ can be calculated akin to Eq.~\eqref{eq:projection} and Eq.~\eqref{eq:residual}.
Then, we remove the semantics already captured by $\mathbf{B}_{t-1}$ in $\mathbf{R}_t^{\mathrm{cand}}$, resulting in $\mathbf{\hat R}_t$:
\begin{equation}
\mathbf{\hat R}_t=\mathbf{R}_t^{\mathrm{cand}}
\left(
\mathbf{I}_d - \mathbf{B}_{t-1}^\top \mathbf{B}_{t-1}
\right),
\end{equation}
where $\mathbf{I}_d \in \mathbb{R}^{d\times d}$ denotes the identity matrix.
We then perform \emph{QR decomposition}\cite{golub2013matrix} on $\mathbf{\hat R}_t$:
\begin{equation}
 \mathbf{\hat R}_t=\mathbf{\bar Q}_t \mathbf{\bar R}_t,
\end{equation}
where $\mathbf{\bar Q}_t \in \mathbb{R}^{k\times{d}}$ provides the orthonormal new basis.
The basis and activity scores are updated as:
\begin{equation}
\mathbf{B}_t^{'}=[\mathbf{B}_{t-1},\mathbf{\bar Q}_t],~~s_t^{'}=[s_{t-1}, {\bar s}_t],
\end{equation}
where $\bar s_t \in \mathbb{R}^{k}$ denotes the activity scores of $\mathbf{\bar Q}_t$. And $\mathbf{B}_t^{'}$ remains row-orthonormal throughout the whole processing, the proof is provided in Appendix ~\ref{alg:causalmem}.

Meanwhile, to adapt to the evolving video stream, we dynamically update the activity score $s_t$. 
Specifically, we denote the $i$-th token of $t$-th frame as $\mathbf{X}_{t,i}$. 
For the $j$-th basis vector of $\mathbf{B}_t$ at the $t$-th time step, we define it as $\mathbf{B}_{t,j}$ and its instantaneous activity as $a_{t,j}$:
\begin{equation}
a_{t,j} = \frac{1}{N} \sum_{i=1}^{N} |\mathbf{B}_{t,j}\mathbf{X}_{t,i}^\top|.
\end{equation}
Then we update the activity score of the $j$-th vector $s_{t,j}$ using an exponential moving average:
\begin{equation}
  s_{t,j} = \gamma s_{t-1,j} + (1-\gamma) a_{t,j},
  \label{eq:basis_activity}
\end{equation}
where $a_{t,j}$ measures the instantaneous activation of the $j$-th basis vector in the current frame, and $s_{t,j}$ denotes the temporally smoothed activity score that smoothes historical activity with recent observations.
Here, $\gamma \in [0,1]$ is the smoothing factor.

Once the number of vectors in $\mathbf{B}_t^{'}$ exceeds the predefined limit $q$, we retain the $q$
 most active basis vectors according to $s_t^{'}$. 
We denote $\mathcal{J}_t$ as the retained index set.
$\mathbf{B}_t$ and $\mathbf{s}_t$ are updated by:
\begin{equation}
\mathbf{B}_t \leftarrow \mathbf{B}_t^{'}(\mathcal{J}_t), \quad
\mathbf{s}_t \leftarrow \mathbf{s}_t^{'}(\mathcal{J}_t),
\quad
\mathrm{where}\quad
\mathcal{J}_t
=
\arg\max_{|\mathcal{J}|=q}
\sum_{j\in\mathcal{J}} s_{t,j}.
\end{equation}
A larger $s_{t,j}$ indicates that the corresponding basis vector has been used more consistently in recent frames.  
Therefore, we retain $q$ most active basis vectors as the updated basis.
In this way, the basis remains a compact yet informative semantic representation with bounded size while evolving online.

\subsection{Dynamic Visual Memory Bank}
As the video stream evolves, the number of visual tokens grows linearly over time. 
To satisfy a fixed token budget, we maintain a dynamic visual memory bank $\mathbf{M}_t \in \mathbb{R}^{b_t\times d}$
at the $t$-th time step. which maximizes the 
information preserved in the video stream.
Here, $b_t\leq b$, $b_t$ is the number of tokens stored in $\mathbf{M}_t$ and $b$ is the fixed budget of the memory bank.
The memory bank $\mathbf{M}_t$ stores a compact set of visual tokens that are
either informative
or temporally recent from the observed video stream and is updated online whenever a new
frame arrives.

At the $t$-th time step, we first insert all tokens in $\mathbf{X}_t$ into the previous memory bank
$\mathbf{M}_{t-1}$, yielding a temporary memory bank
$
\mathbf{M}'_t = [\mathbf{M}_{t-1}, \mathbf{X}_t].
$
If $|\mathbf{M}'_t| > b$, we compress $\mathbf{M}'_t$ guided by the
online semantic basis $\mathbf{B}_t$.

Specifically, for each token $\mathbf{X}_i^{'} \in \mathbf{M}'_t$, we first compute its redundancy score
$\tilde{e}_i$ according to the residual between $\mathbf{X}_i^{'}$ and its projection onto $\mathbf{B}_t$, similar to Eq.(7):
\begin{equation}
  \tilde{e}_i =
\frac{e_i - e_{\min}}{e_{\max} - e_{\min} + \epsilon}, \quad \mathrm{where}\quad
e_i = \left\| x_i - x_i \mathbf{B}_t^\top \mathbf{B}_t  \right\|_2,
\end{equation}
where $e_{\min}$ and $e_{\max}$ are the minimum and maximum redundancy scores among all
tokens in $\mathbf{M}'_t$, respectively, and $\epsilon$ is a small constant to avoid division
by zero. 


In addition to redundancy, we also incorporate a temporal recency prior to encourage the memory
bank to adapt smoothly to the evolving video stream. Therefore, for each token $\mathbf{X}_{i}^{'} \in \mathbf{M}_t^{'}$, we
compute the final retention score as $f_i$ :
\begin{equation}
f_i = (1-\alpha) \tilde{e}_i + \alpha\tilde{\tau}_i,
\quad \mathrm{where}\quad
\tilde{\tau}_i = \frac{t_i + 1}{t + 1},
\label{eq:token_score}
\end{equation}
where $t_i$ denotes the timestamp of the source frame index of token $\mathbf{X}_{i}^{'}$ and
$\alpha$ balances the redundancy score and the temporal recency score.


We then retain $b$ tokens with the highest scores from
$\mathbf{M}'_t$ to form the updated memory bank $\mathbf{M}_t$:
\begin{equation}
\mathbf{M}_t \leftarrow \mathbf{M}^{'}_t(\mathcal{O}_t), \quad \mathrm{where}\quad
\mathcal{O}_t
=
\arg\max_{|\mathcal{O}|=b}
\sum_{i\in\mathcal{O}} f_{i}.
\end{equation}
Tokens that are not selected into $\mathbf{M}_t$ are directly removed. 
In this way,
$\mathbf{M}_t$ always keeps the fixed budget while preserving tokens that
maximize the information of the video stream.

When a question $Q$ arrives at the $t$-th time step, the response $Y$ is generated conditioned on the question
and the visual memory bank $\mathbf{M}_t$:
\begin{equation}
     p(Y \mid \mathbf{M}_t, Q) = \prod_{i=1}^{L} p(y_i \mid G,\mathbf{M}_t, Q, Y_{<i})
\end{equation}

The illustration of CausalMem is given in Fig.~\ref{fig:overall}, and the algorithm is provided in Appendix \ref{alg:causalmem}.

\section{Experiment}
\vspace{-0.1cm}
\subsection{Benchmarks and Metrics}
\vspace{-0.1cm}
We conduct extensive experiments on both streaming and offline video understanding benchmarks to validate the effectiveness of CausalMem. 
For streaming evaluation, we adopt OVO-Bench~\cite{niu2025ovo} and StreamingBench~\cite{lin2026streamingbench}, both of which evaluate real-time video understanding under streaming settings. 
Specifically, OVO-Bench focuses on timestamp-aware streaming reasoning, while StreamingBench emphasizes continuous long-form video comprehension. 
Following prior works, we report results on their real-time subsets.
For offline evaluation, we use Video-MME~\cite{fu2025video}, LongVideoBench~\cite{wu2024longvideobench}, MLVU~\cite{zhou2025mlvu}, and LVBench~\cite{wang2025lvbench}. 
These benchmarks cover a wide range of video durations and reasoning complexities. 
 Video-MME and LongVideoBench contain videos ranging from several seconds to nearly one hour, while MLVU extends the maximum duration to approximately two hours. LVBench further evaluates extreme long-video understanding with videos ranging from 1 to 2 hours.

\vspace{-0.1cm}
\subsection{Implementation} 
\vspace{-0.1cm}
We mainly compare CausalMem with representative streaming video understanding methods, including ReKV~\cite{di2025streaming}, LiveVLM~\cite{ning2025livevlm}, StreamMem~\cite{yang2025streammem}, and StreamingTOM~\cite{chen2025streamingtom}. 
Following their settings, we apply CausalMem to LLaVA-OneVision~\cite{li2024llava} and Qwen2.5-VL~\cite{bai2025qwen25vltechnicalreport}.
For streaming benchmarks, all videos are sampled at 0.5 fps. 
For offline benchmarks, LLaVA-OneVision uses 0.5 fps for videos shorter than 30 minutes and 0.2 fps for longer videos, while Qwen2.5-VL uniformly adopts 1 fps for all videos.
We set the total visual token budget $b$ to 12$k$ for LLaVA-OneVision and 6$k$ for Qwen2.5-VL. 
The maximum number of basis vectors is set to $q=64$, and the maximum number of candidate tokens used for basis update in each frame is set to $m=8$. 
The activity smoothing factor $\gamma$ is set to 0.9. 
In addition, the balancing factor $\alpha$, which controls the trade-off between redundancy and temporal relevance, is set to 0.8 for streaming benchmarks, while the temporal score is disabled for offline evaluation. 
All experiments are conducted on NVIDIA A800 GPUs.

\vspace{-0.1cm}
\subsection{Quantitative Analysis}
\vspace{-0.1cm}
\begin{table*}[t]
  \centering
  \small
  \setlength{\tabcolsep}{4.5pt}
  \renewcommand{\arraystretch}{1.15}
  \caption{Comparison with existing methods on two streaming benchmarks. The base MLLM used is LLaVA-OneVision. The best and second-best results are in \textbf{bold} and \underline{underline} respectively. $\dagger$ denotes the off-line method, which compress the tokens given the whole video frames.}
  \label{tab:results_real_time}
  \resizebox{\linewidth}{!}{
  \begin{tabular}{l c *{18}{c}}
    \toprule
    \multirow{2}{*}{\textbf{Method}} & \multirow{2}{*}{\textbf{Frames}} & \multicolumn{7}{c}{\textbf{OVO-Bench real-time}} & \multicolumn{11}{c}{\textbf{StreamingBench real-time}} \\
    \cmidrule(lr){3-9} \cmidrule(lr){10-20}
    & & \textbf{OCR} & \textbf{ACR} & \textbf{ATR} & \textbf{STU} & \textbf{FPD} & \textbf{OJR} & \textbf{Avg.} 
    & \textbf{OP} & \textbf{CR} & \textbf{CS} & \textbf{ATP} & \textbf{EU} & \textbf{TR} & \textbf{PR} & \textbf{SU} & \textbf{ACP} & \textbf{CT} & \textbf{Avg.} \\
    \midrule

    LLaVA-OneVision~\citep{li2024llava}&32& 67.1 & \textbf{61.5} & 74.1 & 45.5 & 69.3 & 65.2 & 62.6 & 80.4 & 74.2 & 76.0 & \underline{80.7} & 72.7 & 71.7 & 67.6 & 65.5 & 65.7 & 45.1 & 71.1 \\
    \midrule
    VisionZip${\dagger}$~\citep{yang2025visionzip} & 0.5fps & \underline{68.5} & 55.0 & \textbf{77.6} & \underline{48.3} & \underline{74.3} & 65.2 & \underline{63.7} & 79.3 & \textbf{83.6} & 79.7 & \underline{80.3} & 68.4 & 72.1 & 72.2 & 62.4 & \underline{67.7} & 41 & 71.6 \\
    ReKV~\citep{di2025streaming} & 0.5fps & -- & -- & -- & -- & -- & -- & 57.3 & 74.4 & 78.9 & 78.6 & 77.1 & 68.3 & 67.9 & 67.6 & 62.6 & 64.3 & 44.6 & 69.1 \\
    StreamKV~\citep{chen2026streamkv} & 1fps & -- & -- & -- & -- & -- & -- & -- & 74.7 & 78.1 & \textbf{87.7} & 79.4 & \textbf{70.8} & 67.6 & 70.4 & 64.6 & 64.0 & \underline{45.1} & 71.0 \\
    LiveVLM~\citep{ning2025livevlm} & 0.5fps & 67.1 & 60.5 & 70.7 & 46.1 & 71.3 & 61.9 & 61.6 & \underline{81.5} & 78.1 & \underline{83.3} & 79.1 & \underline{69.6} & \underline{74.1} & \textbf{75.0} & \textbf{\underline{69.1}} & \underline{67.7} & 40.4 & \underline{72.9} \\
    StreamingTom~\citep{chen2025streamingtom} & 0.5fps & 67.8 & \textbf{61.5} & 75.9 & 47.8 & 68.3 & \underline{66.3} & 63.6 & 79.2 & \underline{80.1} & 78.2 & 79.4 & 67.7 & 68.9 & 69.4 & 60.0 & 64.6 & \textbf{46.3} & 70.1 \\
    \textbf{CausalMem (Ours)} & 0.5fps & \textbf{71.8} & \textbf{61.5} & \underline{76.7} & \textbf{48.9} & \textbf{76.2} & \textbf{66.8} & \textbf{65.7} & \textbf{82.6} & 79.7 &  83.2 & \textbf{83.2} & \underline{69.6} & \textbf{78.4} & \textbf{75.0} & \underline{67.3} & \textbf{70.9} & 39.4 & \textbf{74.3} \\
    \bottomrule
  \end{tabular}
  }
  \vspace{-0.5cm}
\end{table*}
\textbf{Streaming Benchmarks.} 
We first compare our CausalMem with existing video compression methods on two streaming benchmarks, as reported in Tab. \ref{tab:results_real_time}.
As can be seen, CausalMem consistently improves the baseline models and achieves competitive or superior performance compared with existing methods. 
Specifically, CausalMem improves the baseline from $62.6\%$ to $65.7\%$ on OVO-Bench and from $71.1\%$ to $74.3\%$ on StreamingBench. 
These gains demonstrate that maintaining a compact yet informative visual memory effectively mitigates the unlimited growth of visual tokens in streaming scenarios.
Moreover, CausalMem achieves competitive or even better performance than SOTA streaming methods like StreamMem and StreamingTOM.
For example, on StreamingBench, CausalMem surpasses StreamingTOM and LiveVLM by absolute margins of 4.2\% and 1.4\% in average accuracy, respectively.
This proves the effectiveness of CausalMem that preserves the informative and temporal recent visual tokens in the memory bank.
Furthermore, CausalMem achieves strong performance across multiple subcategories on both streaming benchmarks.
For example, CausalMem obtains the best performance among training-free methods on five out of six OVO-Bench subtasks, \emph{e.g.}, CausalMem outperforms StreamingTOM on the Spatial Understanding and Future Prediction subtasks of OVO-Bench by absolute margins of $1.1\%$ and $7.9\%$, respectively, which shows the robustness of CausalMem on diverse streaming tasks.
In summary, CausalMem is an effective training-free streaming video understanding method that maintains a compact yet informative memory under a fixed budget, thereby helping MLLMs to process the streaming videos.

\begin{table*}[t]
\centering
\caption{Comparison with existing streaming methods on offline benchmarks. The base MLLM used
is LLaVA-OneVision. Streaming methods are evaluated in the streaming setting under strict causality.}
\vspace{-0.15cm}
\resizebox{\textwidth}{!}{
\begin{tabular}{lcccccccccc}

\toprule
\multirow{2}{*}{\textbf{Method}} 
& \multirow{2}{*}{\textbf{Frames}} 
& \multirow{2}{*}{\textbf{Tokens}} 
& \multirow{2}{*}{\textbf{VideoMME}} 
& \multicolumn{4}{c}{\textbf{MLVU}} 
& \multirow{2}{*}{\makecell{\textbf{LongVideo} \\ \textbf{Bench}}} 
& \multirow{2}{*}{\textbf{LVBench}} 
& \multirow{2}{*}{\textbf{Avg}} \\
\cmidrule(lr){5-8}
&  &  &  
& \textbf{Single detail} 
& \textbf{Multi detail} 
& \textbf{Holistic} 
& \textbf{Avg} 
&  &  &  \\

\midrule
LLaVA-OneVision~\citep{li2024llava} 
& 32 
& 6k 
& 58.5 
& 68.6 
& 39.1 
& 79.3 
& 64.7 
& \underline{56.5} 
& 38.4 
& 54.5 \\

\toprule
ReKV~\citep{di2025streaming} 
& 0.5fps 
& 12k 
& 58.3 
& - 
& - 
& - 
& \underline{68.5} 
& 55.8 
& \textbf{47.8} 
& - \\

LiveVLM~\citep{ning2025livevlm} 
& 0.5/0.2fps 
& 12k 
& 59.6 
& - 
& - 
& - 
& 68.1 
& 56.1 
& - 
& - \\

StreamMem~\citep{yang2025streammem} 
& 0.5/0.2fps 
& 6k 
& 59.4 
& - 
& - 
& - 
& 66.9 
& 54.4 
& - 
& - \\

StreamingTOM~\citep{chen2025streamingtom} 
& 0.5/0.2fps 
& 12k 
& \underline{59.9} 
& - 
& - 
& - 
& 67.9 
& 56.4 
& 40.5 
& \underline{56.2} \\

\textbf{CausalMem(Ours)} 
& 0.5/0.2fps 
& 12k 
& \textbf{60.0} 
& \textbf{74.0} 
& \textbf{49.5} 
& \textbf{83.6} 
& \textbf{70.9} 
& \textbf{57.0} 
& \underline{42.1} 
& \textbf{57.5} \\

\bottomrule
\end{tabular}
}

\label{tab:main_results}
\vspace{-0.25cm}
\end{table*}
\begin{table*}[t]
  \centering
  \small
  \setlength{\tabcolsep}{3.8pt}
  \renewcommand{\arraystretch}{1.15}
  \caption{Comparison with advanced MLLMs on two streaming benchmarks.}
  \vspace{-0.15cm}
  \label{tab:sota}
  \resizebox{\linewidth}{!}{
  \begin{tabular}{l c c ccccccccccc}
    \toprule
    \multirow{2}{*}{\textbf{Method}} 
    & \multirow{2}{*}{\textbf{Frames}} 
    & \multirow{2}{*}{\textbf{OVO-Bench}} 
    & \multicolumn{11}{c}{\textbf{StreamingBench}} \\
    
    \cmidrule(lr){4-14}
    
    & & 
    & \textbf{OP} 
    & \textbf{CR} 
    & \textbf{CS} 
    & \textbf{ATP} 
    & \textbf{EU} 
    & \textbf{TR} 
    & \textbf{PR} 
    & \textbf{SU} 
    & \textbf{ACP} 
    & \textbf{CT} 
    & \textbf{Avg} \\
    
    \midrule
    \rowcolor{gray!10}
    Gemini 1.5 Pro~\citep{team2024gemini} & -- & 69.3
    & 79.0 & 80.5 & 83.5 & 79.7 & 80.0 & 84.7 & 77.8 & 64.2 & 72.0 & 48.7 & 75.7 \\
    \rowcolor{gray!10}
    
    GPT-4o~\citep{hurst2024gpt} & -- & 64.5 
    & 77.1 &80.5& 83.9 & 76.5 & 70.2 & 83.8 & 66.7 & 62.2 & 69.1 & 49.2 & 73.3 \\
    \midrule
    LongVA~\citep{zhang2024long} & 128 & -- 
    & 70.0 & 63.3 & 61.2 & 70.9 & 62.7 & 59.5 & 61.1 & 53.7 & 54.7 & 34.7 & 60.0 \\

    LongVU~\citep{shen2024longvu} & 1fps & 57.4 
    & -- & -- & -- & -- & -- & -- & -- & -- & -- & -- & -- \\

    LLaVA-Video~\citep{zhang2024llava} & 64 & 63.5 
    & -- & -- & -- & -- & -- & -- & -- & -- & -- & -- & -- \\

    VideoLLM-Online~\citep{chen2024videollm} & 2fps & 20.8 
    & 39.1 & 40.1 & 34.5 & 31.1 & 46.0 & 32.4 & 31.5 & 34.2 & 42.5 & 27.9 & 36.0 \\

    Dispider~\citep{qian2025dispider} & 1fps & 54.6 
    & 74.9 & 75.5 & 74.1 & 73.1 & 74.4 & 59.9 & 76.1 & 62.9 & 62.2 & 45.8 & 67.6 \\

    Flash-VStream~\citep{zhang2024flash} & 1fps & 28.4 
    & 25.9 & 43.6 & 24.9 & 23.9 & 27.3 & 13.1 & 18.5 & 25.2 & 23.9 & 48.7 & 23.2 \\

    ViSpeak~\citep{fu2025vispeak} & 1fps & \underline{66.3} 
    & 79.8 & \textbf{88.3} & \underline{83.3} & 81.1 & \underline{76.4} & 75.1 & 70.4 & 65.9 & \textbf{77.3} & 34.2 & 74.4 \\

    TimeChat-Online~\citep{yao2025timechat} & 1fps & 61.4 
    & \underline{80.8} & 79.7 & 80.8 & \underline{83.3} & 74.8 & 78.8 & \underline{78.7} & 64.2 & 68.8 & \textbf{58.0} & \underline{75.3} \\


    \midrule
    LLaVA-OneVision~\citep{li2024llava} & 32 & 62.6 
    & 79.3 & 76.6 & 76.2 & 81.3 & 75.9 & 71.5 & 71.3 & 66.9 & 63.4 & 43.1 & 71.3 \\
    
    \textbf{+CausalMem(Ours)} & 0.5/0.2fps & 65.7 
    & \textbf{82.6} & 79.7 & 83.2 & 83.2 & 69.6 & 78.4 & 75.0 & \underline{67.3} & \underline{70.9} & 39.4 & 74.3\\

    Qwen2.5-VL~\citep{bai2025qwen25vltechnicalreport} & 1fps & -- 
    & 78.3 & \underline{80.5} & 79.8 & 82.4 & 75.5 & \underline{80.4} & 74.1 & 62.6 & 67.6 & \underline{51.1} & 73.9 \\

    \textbf{+CausalMem(Ours)} & 1fps & \textbf{67.8}
    & 80.7 & 78.9 & \textbf{87.3} & \textbf{85.5} & \textbf{79.1} & \textbf{84.0} & \textbf{81.5} & \textbf{70.2} & 69.4 & 43.0 & \textbf{76.9} \\

\bottomrule
  \end{tabular}
  }
  \vspace{-0.5cm}
\end{table*}
\textbf{Offline Benchmarks.} Apart from streaming evaluation, we also evaluate CausalMem using LLaVA-OneVision on offline benchmarks
compared with
both streaming and offline methods in Tab. \ref{tab:main_results}. 
For offline evaluation, CausalMem still follows the streaming processing setting under strict causality, where frames are unknown to future information.
The results show that CausalMem also achieves the competitive or even better performance on four
offline benchmarks.
For instance, CausalMem surpasses StreamingTOM by  $3.0\%$ and 
$1.6\%$ absolutely on MLVU and LVBench, respectively.
It demonstrates the robustness of CausalMem across benchmarks targeting 
different scenarios, \emph{e.g.}, comprehensive understanding of MLVU and detailed understanding of LVBench.
It further proves that CausalMem maximizes the diversified information of the video stream in the visual memory bank, which makes it perform well in different scenarios. Overall, these results show that CausalMem is fully capable of both streaming and offline settings.

\textbf{Comparison with SOTA Video-MLLMs.} We further apply CausalMem to LLaVA-OneVision and Qwen2.5-VL, and compare with existing advanced Video-MLLMs on two streaming benchmarks in Tab. ~\ref{tab:sota}.
We can observe that CausalMem achieves highly competitive performance compared with existing advanced Video-MLLMs, even though it is a plug-and-play method. For instance, CausalMem help Qwen2.5-VL to achieve $67.8\%$ on OVO-Bench, outperforming TimeChat-Online by $6.4\%$ absolutely.
It shows that CausalMem effectively strengthens the streaming video understanding ability of MLLMs via maintaining a compact and dynamically updated visual memory bank.
Furthermore, CausalMem improves Qwen2.5-VL to the level of Gemini 1.5 Pro and GPT-4o, \emph{e.g.}, surpassing GPT-4o by $3.3\%$ and $3.6\%$ on OVO-Bench and StreamingBench, respectively.



\begin{figure}[!t]
\vspace{-0.2cm}
  \centering
  \includegraphics[width=0.95\columnwidth]{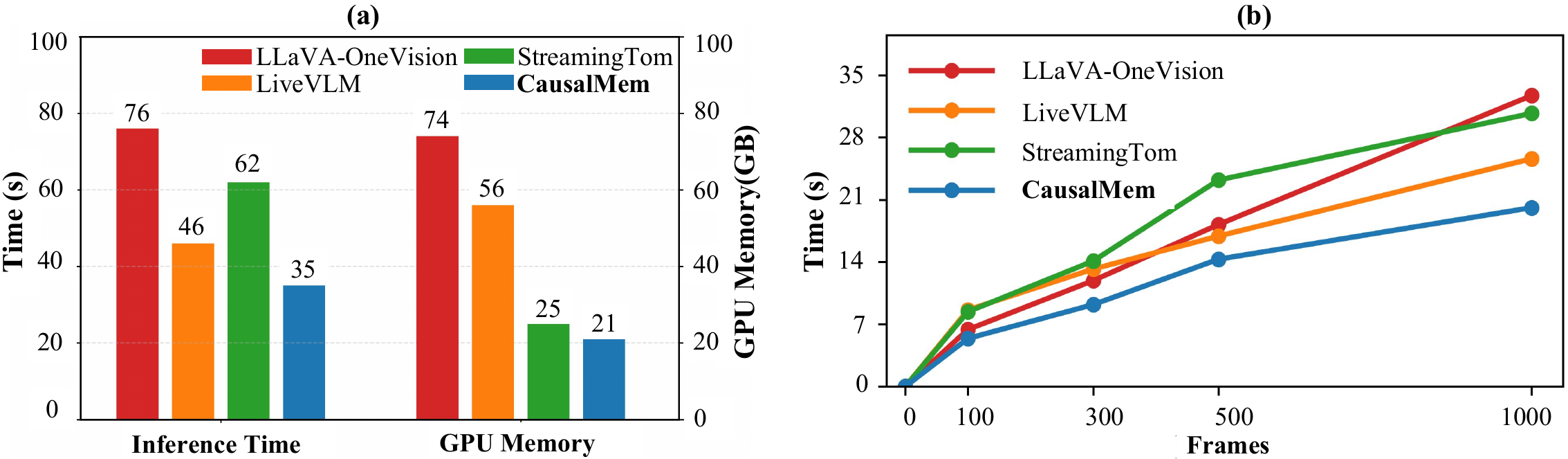}
  \caption{(a) Comparison with existing methods in inference time and GPU memory overhead. (b) The inference time varies with the number of input frames.}
  \label{fig:effience}
  \vspace{-0.2cm}
\end{figure}

\textbf{Analysis of Efficiency.} 
We compare inference time, GPU memory usage, and scalability with existing streaming methods in Fig.\ref{fig:effience}.
Following the online setting, a 30-minute video is sampled at 1 FPS and processed frame by frame. GPU memory denotes the peak memory during inference, and the inference time is computed from the first frame entering the ViT to the generation of the first token by LLM.
As shown in Fig.\ref{fig:effience}-(a), CausalMem consumes both the lowest GPU and inference time overhead. This advantage mainly stems from the fixed-budget memory design. Unlike LLaVA-OneVision, whose visual tokens grow linearly with incoming frames, CausalMem keeps the visual memory bank fixed and feeds a compact token sequence to MLLM. Moreover, CausalMem avoids the extra overhead introduced by KV-cache retrieval or quantization in existing streaming methods.
Fig.\ref{fig:effience}-(b) further shows that CausalMem scales better as the number of frames increases, exhibiting the slowest growth in inference time compared with other methods. These results demonstrate that the fixed-budget memory bank effectively reduces both memory usage and inference time for long streaming videos.

\begin{table*}[t]
\centering
\caption{Ablation studies of key components in CausalMem. $*$ is the final setting of CausalMem.}
\vspace{-0.1cm}
\label{tab:ablation_all}

\footnotesize
\renewcommand{\arraystretch}{1.12}
\setlength{\tabcolsep}{3pt}

\begin{minipage}[t]{0.47\textwidth}
\centering
{\footnotesize \textbf{(a) Basis Size $q$}}\\
\begin{tabular*}{\linewidth}{@{\extracolsep{\fill}}lccc@{}}
\toprule
\textbf{$q$} & \makecell[c]{\textbf{Streaming}\\\textbf{Bench}} & \makecell[c]{\textbf{OVO-}\\\textbf{Bench}} & \textbf{MLVU} \\
\midrule
0   & 74.0 & 65.3 & 68.9 \\
16  & 74.0 & 65.5 & 70.7 \\
64$*$  & \textbf{74.3} & \textbf{65.7} & \textbf{70.9} \\
512 & 73.7 & 65.2 & 70.1 \\
\bottomrule
\end{tabular*}
\end{minipage}
\hfill
\begin{minipage}[t]{0.47\textwidth}
\centering
{\footnotesize\bfseries (b) Basis Update Tokens $m$}\\
\begin{tabular*}{\linewidth}{@{\extracolsep{\fill}}lccc@{}}
\toprule
\textbf{$m$} & \makecell[c]{\textbf{Streaming}\\\textbf{Bench}} & \makecell[c]{\textbf{OVO-}\\\textbf{Bench}} & \textbf{MLVU} \\
\midrule
4  & 74.1 & 65.5 & 70.7 \\
8$*$  & \textbf{74.3} & \textbf{65.7} & \textbf{70.9} \\
16 & \textbf{74.3}   & 65.6   & 70.8  \\
32 & 74.0 & 65.1 & 70.4 \\
\bottomrule
\end{tabular*}
\end{minipage}

\vspace{0.55em}

\begin{minipage}[t]{0.47\textwidth}
\centering
{\footnotesize\bfseries (c) Basis Activity Score $s_{t,j}$}\\
\begin{tabular*}{\linewidth}{@{\extracolsep{\fill}}lccc@{}}
\toprule
\textbf{Setting} & \makecell[c]{\textbf{Streaming}\\\textbf{Bench}} & \makecell[c]{\textbf{OVO-}\\\textbf{Bench}} & \textbf{MLVU} \\
\midrule
only Historical Score & 74.1 & 65.3 & 70.4 \\
only Instantaneous Score  & 74.0 & 65.2 & 70.6 \\
Both$*$   & \textbf{74.3} & \textbf{65.7} & \textbf{70.9} \\
\bottomrule
\end{tabular*}
\end{minipage}
\hfill
\begin{minipage}[t]{0.47\textwidth}
\centering
{\footnotesize\bfseries (d) Token Retention Score $f_i$}\\
\begin{tabular*}{\linewidth}{@{\extracolsep{\fill}}lccc@{}}
\toprule
\textbf{Setting} & \makecell[c]{\textbf{Streaming}\\\textbf{Bench}} & \makecell[c]{\textbf{OVO-}\\\textbf{Bench}} & \textbf{MLVU} \\
\midrule
only Redundancy score & 74.0 & 65.3 & 60.2 \\
only Temporal score & 71.8 & 65.1 & 70.9 \\
Both$*$   & \textbf{74.3} & \textbf{65.7} & \textbf{71.1} \\
\bottomrule
\end{tabular*}
\end{minipage}
\vspace{-0.4cm}
\end{table*}

\textbf{Ablation Study.} We conduct ablation studies 
of the key designs of CausalMem in Tab. \ref{tab:ablation_all}. The four subtables respectively correspond to (a) the basis size $q$ of the online semantic basis, (b) the number of tokens $m$ used to update the basis, (c) the basis activity score $s_{t,j}$ in Eq.\eqref{eq:basis_activity}, and (d) the token retention score\eqref{eq:token_score}.
We first ablate the number of basis vectors $q$ in Tab. \ref{tab:ablation_all} (a). Note that $q=0$ denotes that we compute the redundancy of visual tokens upon the whole memory bank, instead of the online semantic basis.
As shown, online benchmarks decreased slightly without the online semantic basis, due to preserving the temporally recent tokens. 
However, on the offline benchmark MLVU, the performance drops sharply by $2.0\%$.
It demonstrates the suboptimality of using the whole memory bank to estimate the token redundancy and the effectiveness of the online semantic basis.
Then we can observe that the performance improves as $q$ increases from 16 to 64, 
then gradually decreases with larger $q$. 
It shows that a small basis may underrepresent the principal semantics of the video stream, leading to inaccurate redundancy estimation, while an overly large basis may absorb noisy or overly detailed information, and thus weaken the discriminativeness of the token redundancy. 
Therefore, adopt a moderate value of $q=64$ to balance semantic coverage and the discriminative capability of redundancy estimation.
Second, we ablate the number of tokens 
$m$ used to update the online semantic basis per frame in Tab.\ref{tab:ablation_all}-(b). 
As shown, the performance first slightly improves when increasing $m$ from 4 to 8, and then gradually decreases with larger $m$. 
This suggests that a moderate number of tokens provides sufficient new semantics for basis update, while too few tokens may make it difficult to cover the new semantics in a new frame, and too many tokens may introduce noisy or overly fine-grained information. 

Furthermore, we analyze the effect of the   \emph{activity score} of the semantic basis in Eq.~\eqref{eq:basis_activity} From Tab. \ref{tab:ablation_all}-(c). We can observe that removing either score leads to suboptimal performance.
This suggests that relying only on historical activity makes the semantic basis difficult to adapt to the evolving video stream, while using only the instantaneous activity of the current frame may cause the basis to rapidly forget historical semantics. In contrast, combining historical activity with current-frame instantaneous activity enables the basis to adapt to new semantics while preserving previously observed semantics, leading to the best performance.
Besides, we analyze the effect of Token Retention Score in Eq.~\eqref{eq:token_score}, as reported in Tab. \ref{tab:ablation_all}-(d).
Without the redundancy score, CausalMem still gains competitive performance on streaming benchmarks, since the temporal score helps preserve recently observed tokens. However, its performance drops significantly on the offline benchmark MLVU, showing that the redundancy score is crucial for maintaining globally informative and semantically diverse tokens. 
In contrast, removing the temporal score weakens the performance on streaming benchmarks, suggesting that temporal recency is important for the streaming scenario. By combining both scores, CausalMem achieves the best overall performance, demonstrating its complementary effects in maintaining an informative and temporal recency memory bank.

\vspace{-0.2cm}
\subsection{Qualitative Analysis}
\begin{figure}[t]
    \centering
    \includegraphics[width=0.95\linewidth]{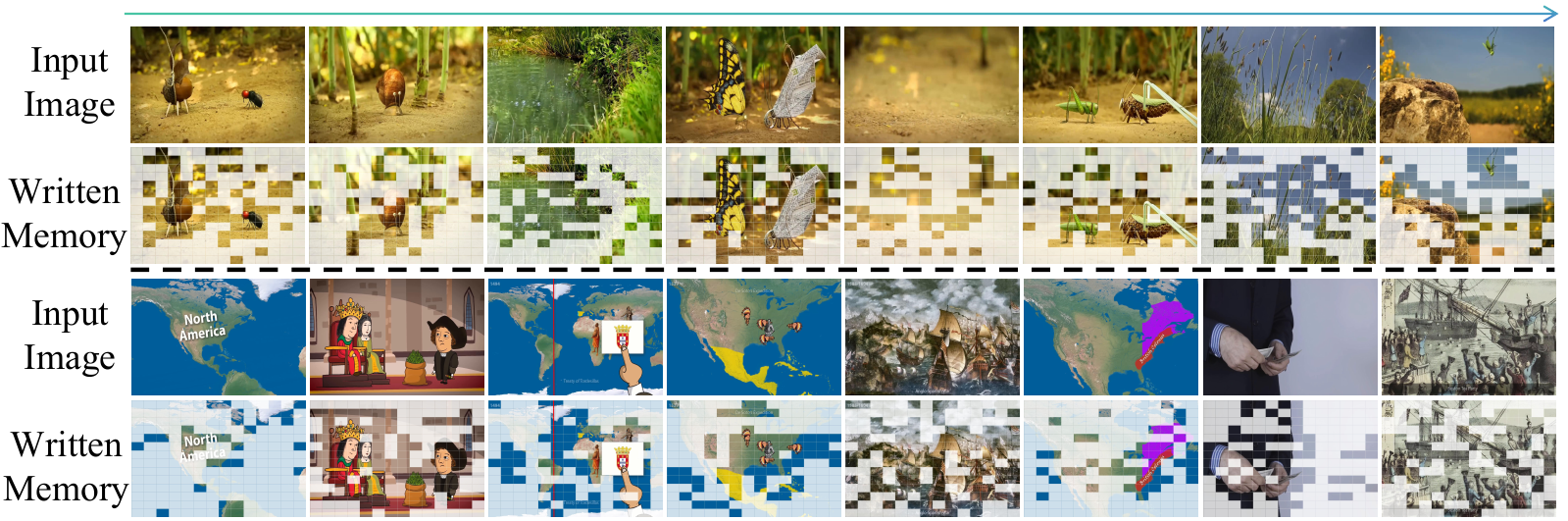}
    \caption{Visualizations of CausalMem. \emph{Input image} refers to the streaming video frames, and \emph{Written Memory} denotes the visual tokens written into the memory bank. These results show that CausalMem can maintain an informative and compact memory bank under a fixed token budget for streaming videos.}
    \label{fig:example}
    \vspace{-0.4cm}
\end{figure}
Fig.~\ref{fig:example} visualizes how CausalMem maintains the visual memory bank during streaming video processing. The ``Input Image'' rows present the incoming video frames, while the ``Written Memory'' rows show the visual tokens selected and written into the memory bank at each time step.
As the video stream evolves, CausalMem adaptively preserves informative visual content instead of uniformly retaining tokens from all frames. 
Specifically, redundant background regions and repetitive visual patterns are largely discarded under a fixed memory budget, whereas tokens corresponding to salient objects, distinctive scene regions, and newly emerging semantic content are consistently preserved. 
These results demonstrate that CausalMem can effectively reduce redundancy tokens while preserving informative tokens, enabling the model to maintain a compact yet informative visual memory throughout streaming video processing.
\section{Conclusion}
In this paper, we study the efficient streaming video understanding from the perspective of a dynamic and fixed-budget visual memory bank, and propose a novel and training-free approach for MLLMs, termed CausalMem. To build such a visual memory bank, CausalMem introduces an online and evolving semantic basis to estimate the principal semantics of streaming tokens, based on which the memory tokens in the bank are written and updated. Extensive experiments on two representative MLLMs and a set of streaming and offline benchmarks show its advantages to existing methods, and also confirm its superiority in semantic preservation for streaming videos.

\bibliographystyle{unsrtnat}
\bibliography{references}








\appendix

\section{ Limitations}
CausalMem is a training-free method that applies on existing MLLMs by maintaining a dynamical fixed-budget visual memory, without additional training or fine-tuning. Therefore, its performance is naturally influenced by the capability of the backbone MLLM, including its visual perception, temporal reasoning, and instruction-following abilities. When applied to stronger MLLMs, CausalMem may further benefit from their improved video understanding capability. Exploring its adaptation to more advanced MLLMs is our future work.
\section{Orthonormality of the Online Semantic Basis.}
We represent the online semantic basis as
\(\mathbf{B}_t\in\mathbb{R}^{q_t\times d}\), where each row is a semantic
basis vector. We show that the update preserves the row-orthonormality of
\(\mathbf{B}_t\), i.e.,
\[
\mathbf{B}_t\mathbf{B}_t^\top=\mathbf{I}_{q_t}.
\]

At initialization, \(\mathbf{B}_1\) is constructed from the retained right
singular vectors of \(\mathbf{X}_1\). Since right singular vectors are
orthonormal, we have
\[
\mathbf{B}_1\mathbf{B}_1^\top=\mathbf{I}_{q_1}.
\]

Assume that at time step \(t-1\), the basis is row-orthonormal:
\[
\mathbf{B}_{t-1}\mathbf{B}_{t-1}^{\top}
=
\mathbf{I}_{q_{t-1}}.
\]
Then
\[
\mathbf{P}_{t-1}
=
\mathbf{B}_{t-1}^{\top}\mathbf{B}_{t-1}
\]
is the orthogonal projector onto the subspace spanned by the rows of
\(\mathbf{B}_{t-1}\). Therefore,
\[
\mathbf{P}_{t-1}^{\perp}
=
\mathbf{I}_d-\mathbf{B}_{t-1}^{\top}\mathbf{B}_{t-1}
\]
is the projector onto its orthogonal complement.

For the candidate residual matrix \(\mathbf{R}^{cand}_t\), we remove the
components already represented by the current basis:
\[
\hat{\mathbf{R}}_t
=
\mathbf{R}^{cand}_t
\left(
\mathbf{I}_d-\mathbf{B}_{t-1}^{\top}\mathbf{B}_{t-1}
\right).
\]
Thus,
\[
\begin{aligned}
\hat{\mathbf{R}}_t\mathbf{B}_{t-1}^{\top}
&=
\mathbf{R}^{cand}_t
\left(
\mathbf{I}_d-\mathbf{B}_{t-1}^{\top}\mathbf{B}_{t-1}
\right)
\mathbf{B}_{t-1}^{\top} \\
&=
\mathbf{R}^{cand}_t\mathbf{B}_{t-1}^{\top}
-
\mathbf{R}^{cand}_t
\mathbf{B}_{t-1}^{\top}
\mathbf{B}_{t-1}
\mathbf{B}_{t-1}^{\top} \\
&=
\mathbf{R}^{cand}_t\mathbf{B}_{t-1}^{\top}
-
\mathbf{R}^{cand}_t
\mathbf{B}_{t-1}^{\top}
\left(
\mathbf{B}_{t-1}\mathbf{B}_{t-1}^{\top}
\right) \\
&=
\mathbf{R}^{cand}_t\mathbf{B}_{t-1}^{\top}
-
\mathbf{R}^{cand}_t\mathbf{B}_{t-1}^{\top}
=
\mathbf{0}.
\end{aligned}
\]
Hence, every row of \(\hat{\mathbf{R}}_t\) lies in the orthogonal complement
of the current basis.

We then apply QR orthonormalization to obtain new row-orthonormal basis
vectors \(\bar{\mathbf{Q}}_t\in\mathbb{R}^{r_t\times d}\), where
\(r_t\leq k\), such that
\[
\bar{\mathbf{Q}}_t\bar{\mathbf{Q}}_t^\top=\mathbf{I}_{r_t}.
\]
Since the rows of \(\bar{\mathbf{Q}}_t\) are linear combinations of the rows
of \(\hat{\mathbf{R}}_t\), they remain orthogonal to the previous basis:
\[
\bar{\mathbf{Q}}_t\mathbf{B}_{t-1}^{\top}=\mathbf{0}.
\]

The updated basis before pruning is obtained by concatenating along the
basis-vector dimension:
\[
\mathbf{B}'_t
=
\begin{bmatrix}
\mathbf{B}_{t-1}\\
\bar{\mathbf{Q}}_t
\end{bmatrix}.
\]
Therefore,
\[
\begin{aligned}
\mathbf{B}'_t{\mathbf{B}'_t}^{\top}
&=
\begin{bmatrix}
\mathbf{B}_{t-1}\\
\bar{\mathbf{Q}}_t
\end{bmatrix}
\begin{bmatrix}
\mathbf{B}_{t-1}^{\top} & \bar{\mathbf{Q}}_t^{\top}
\end{bmatrix} \\
&=
\begin{bmatrix}
\mathbf{B}_{t-1}\mathbf{B}_{t-1}^{\top}
&
\mathbf{B}_{t-1}\bar{\mathbf{Q}}_t^{\top}
\\
\bar{\mathbf{Q}}_t\mathbf{B}_{t-1}^{\top}
&
\bar{\mathbf{Q}}_t\bar{\mathbf{Q}}_t^{\top}
\end{bmatrix} \\
&=
\begin{bmatrix}
\mathbf{I}_{q_{t-1}} & \mathbf{0}\\
\mathbf{0} & \mathbf{I}_{r_t}
\end{bmatrix}.
\end{aligned}
\]
Thus, \(\mathbf{B}'_t\) is row-orthonormal. If the number of basis vectors
exceeds the budget \(q\), we retain a subset of rows according to the activity
scores. Since any subset of a row-orthonormal matrix remains row-orthonormal,
the pruned basis \(\mathbf{B}_t\) also satisfies
\[
\mathbf{B}_t\mathbf{B}_t^\top=\mathbf{I}_{q_t}.
\]
By induction, the online semantic basis remains row-orthonormal at every time
step.

\section{Additional Visualization Results}

We provide qualitative visualizations to illustrate how the proposed method processes long video streams and maintains informative visual content.

The results show that our method effectively preserves informative and semantically salient content while maintaining a compact representation of the video stream.
\clearpage

\begin{figure}[H]
\centering

\begin{subfigure}{1\linewidth}
    \centering
    \includegraphics[width=\linewidth]{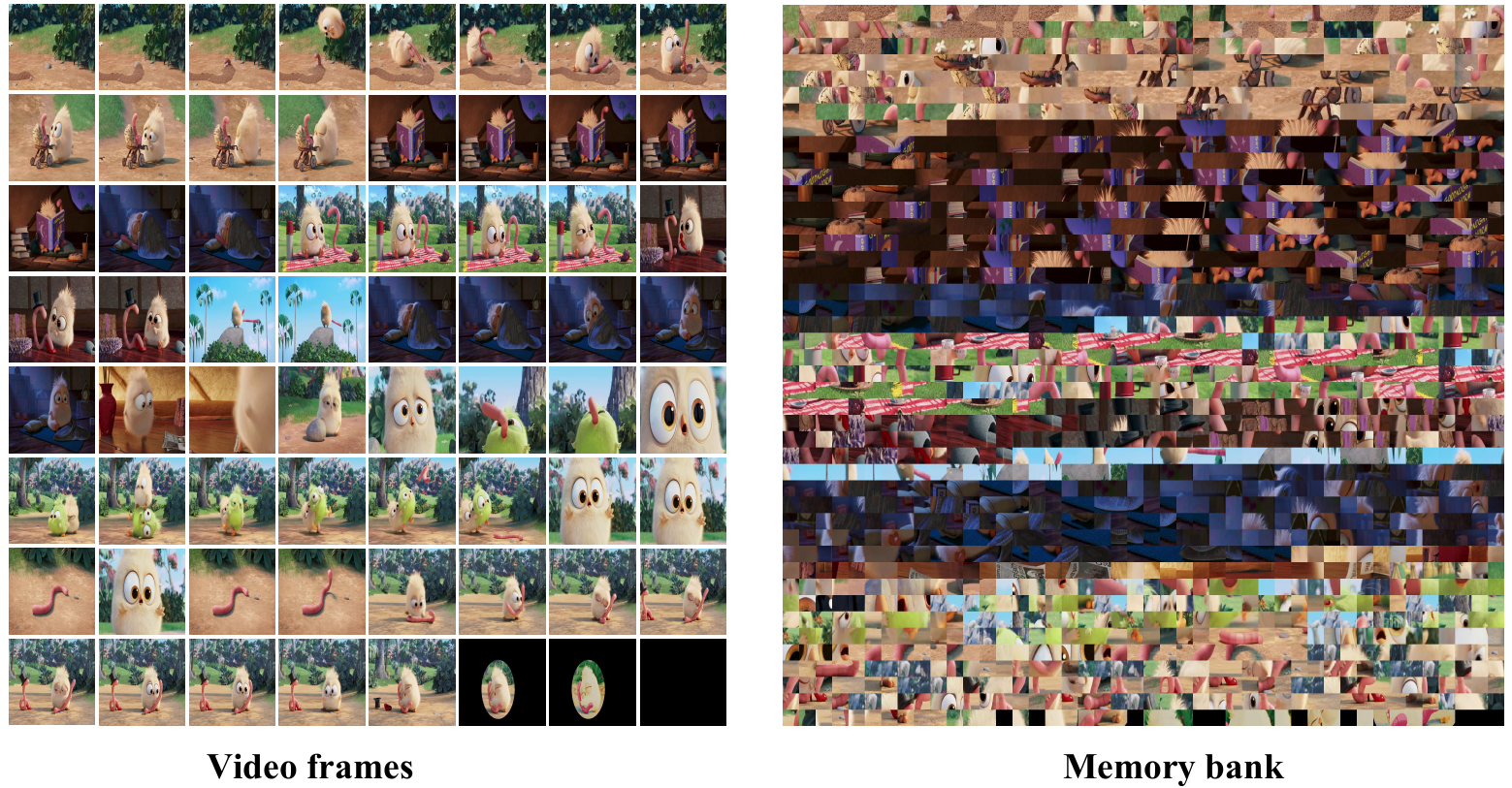}
    \caption{Video 1}
\end{subfigure}

\vspace{0.3cm}

\begin{subfigure}{1\linewidth}
    \centering
    \includegraphics[width=\linewidth]{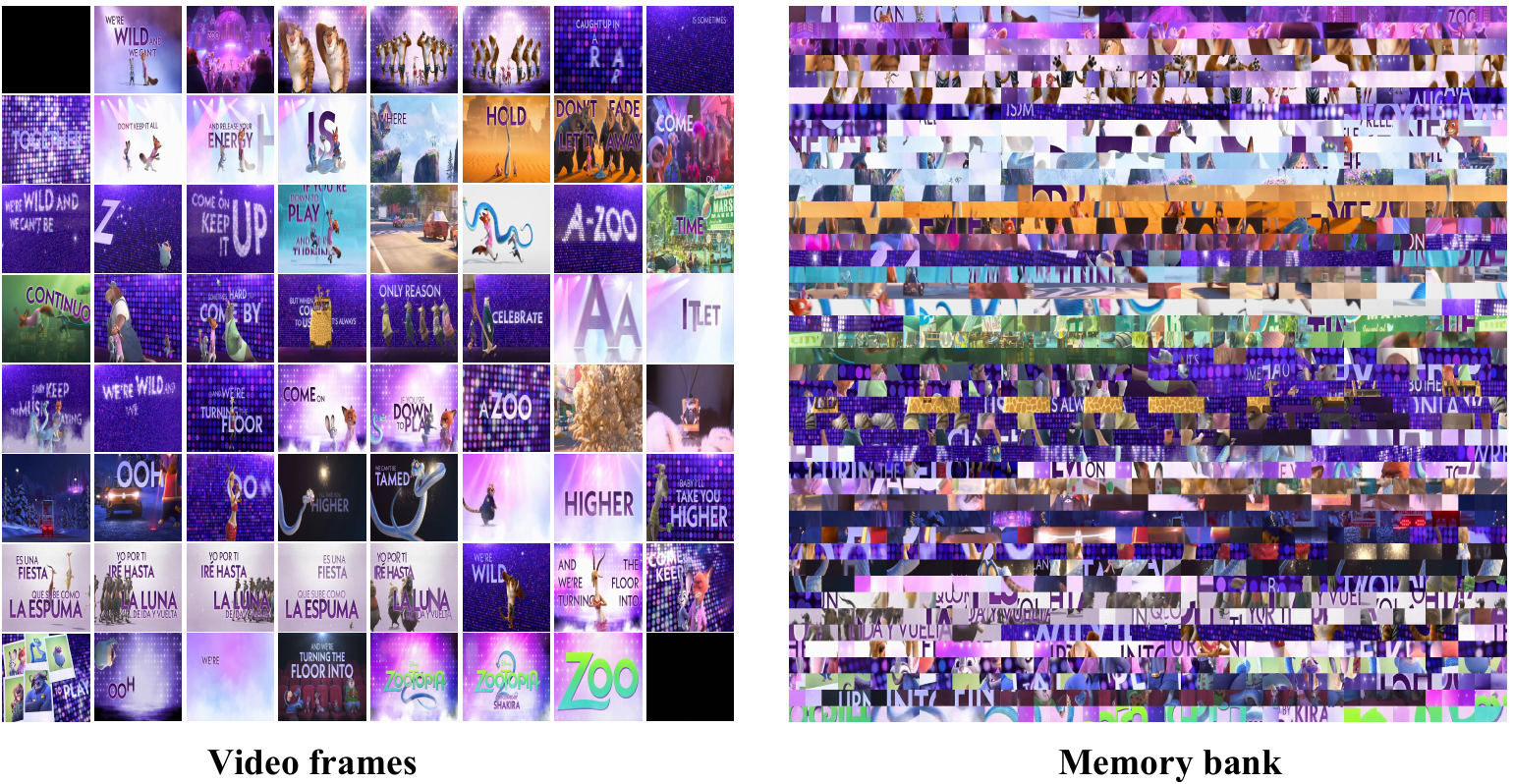}
    \caption{Video 2}
\end{subfigure}

\caption{Qualitative visualization of raw video frames and the corresponding memory bank representations. }
\label{fig:frame_vs_memory_vertical}
\end{figure}
\section{ Algorithm}
\clearpage
\begin{algorithm}[t]
\caption{CausalMem$(\cdot)$}
\label{alg:causalmem}
\begin{algorithmic}[1]
\Require Streaming video $\mathcal{V}=\{I_t\}_{t=1}^{T}$, question $Q$, MLLM generator $G$, memory budget $b$, basis budget $q$, basis update size $m$, balance factor $\alpha$, decay factor $\gamma$
\Ensure Answer $Y$, memory bank $\mathbf{M}_T$

\State $\mathbf{M}_0 \leftarrow \emptyset$, $\mathbf{B}_0 \leftarrow \emptyset$, $\mathbf{s}_0 \leftarrow \emptyset$

\For{$t=1$ to $T$}
    \State $\mathbf{X}_t \leftarrow \Phi(I_t)$
    
    \If{$\mathbf{B}_{t-1}=\emptyset$}
        \State $\mathbf{B}_t \leftarrow \textsc{TopSVD}(\mathbf{X}_t,q_1)$, \quad $\mathbf{s}_t \leftarrow \mathbf{1}_{q_1}$
    \Else
        \State $(\mathbf{R}_t,\mathbf{e}_t) \leftarrow \textsc{ResidualScore}(\mathbf{X}_t,\mathbf{B}_{t-1})$
        \State $\mathbf{X}_t^{\mathrm{cand}} \leftarrow \textsc{TopK}(\mathbf{X}_t,\mathbf{e}_t,m)$
        \State $\bar{\mathbf{Q}}_t \leftarrow \textsc{Orthogonalize}(\mathbf{X}_t^{\mathrm{cand}},\mathbf{B}_{t-1})$
        \State $\mathbf{B}'_t \leftarrow [\mathbf{B}_{t-1};\bar{\mathbf{Q}}_t]$, \quad $\mathbf{s}'_t \leftarrow [\mathbf{s}_{t-1};\mathbf{1}]$
        \State $\mathbf{s}'_t \leftarrow \textsc{UpdateActivity}(\mathbf{s}'_t,\mathbf{B}'_t,\mathbf{X}_t,\gamma)$
        \State $(\mathbf{B}_t,\mathbf{s}_t) \leftarrow \textsc{KeepTopBasis}(\mathbf{B}'_t,\mathbf{s}'_t,q)$
    \EndIf

    \State $\mathbf{M}'_t \leftarrow [\mathbf{M}_{t-1};\mathbf{X}_t]$

    \If{$|\mathbf{M}'_t|>b$}
        \State $\mathbf{f}_t \leftarrow \textsc{RetentionScore}(\mathbf{M}'_t,\mathbf{B}_t,\alpha,t)$
        \State $\mathbf{M}_t \leftarrow \textsc{TopK}(\mathbf{M}'_t,\mathbf{f}_t,b)$
    \Else
        \State $\mathbf{M}_t \leftarrow \mathbf{M}'_t$
    \EndIf
\EndFor

\State $Y \sim G(\mathbf{M}_T,Q)$
\State \Return $Y,\mathbf{M}_T$

\end{algorithmic}
\end{algorithm}

\clearpage

\end{document}